\documentclass[lettersize,journal]{IEEEtran}
\usepackage{amsmath,amsfonts}
\usepackage{array}
\usepackage[caption=false,font=normalsize,labelfont=sf,textfont=sf]{subfig}
\usepackage{textcomp}
\usepackage{stfloats}
\usepackage{url}
\usepackage{verbatim}
\usepackage{graphicx}
\def\BibTeX{{\rm B\kern-.05em{\sc i\kern-.025em b}\kern-.08em
    T\kern-.1667em\lower.7ex\hbox{E}\kern-.125emX}}
\usepackage{balance}

\usepackage{cite}
\usepackage{amssymb}
\usepackage{graphicx}
\usepackage{float}
\usepackage{placeins}
\usepackage{booktabs}
\usepackage{multirow}
\usepackage{tcolorbox}
\usepackage{algorithm}
\usepackage{algpseudocode}
\usepackage{enumitem}
\usepackage{amsthm}
\usepackage{hyperref}
\usepackage{xspace}

\begin{document}
\title{Knowledge Pyramid Construction for Multi-Level Retrieval-Augmented Generation}
\author{
    \IEEEauthorblockN{
        Rubing Chen\IEEEauthorrefmark{2},
        Xulu Zhang\IEEEauthorrefmark{2},
        Jiaxin Wu\IEEEauthorrefmark{2},
        Wenqi Fan\IEEEauthorrefmark{2},
        Xiao-Yong Wei\IEEEauthorrefmark{2},
        Qing Li\IEEEauthorrefmark{2}\IEEEauthorrefmark{1} \\
        \IEEEauthorblockA{
        \IEEEauthorrefmark{2}The Hong Kong Polytechnic University
        }
    }
}



\maketitle

\begin{abstract}

This paper addresses the need for improved precision in existing knowledge-enhanced question-answering frameworks, specifically Retrieval-Augmented Generation (RAG) methods that primarily focus on enhancing recall. 
We propose a multi-layer knowledge pyramid approach within the RAG framework to achieve a better balance between precision and recall. 
The knowledge pyramid consists of three layers: Ontologies, Knowledge Graphs (KGs), and chunk-based raw text. 
We employ cross-layer augmentation techniques for comprehensive knowledge coverage and dynamic updates of the Ontology schema and instances. 
To ensure compactness, we utilize cross-layer filtering methods for knowledge condensation in KGs. 
Our approach, named PolyRAG, follows a waterfall model for retrieval, starting from the top of the pyramid and progressing down until a confident answer is obtained. 
We introduce two benchmarks for domain-specific knowledge retrieval, one in the academic domain and the other in the financial domain. 
The effectiveness of the methods has been validated through comprehensive experiments by outperforming 19 SOTA methods.
An encouraging observation is that the proposed method has augmented the GPT-4, providing 395\% F1 gain by improving its performance from 0.1636 to 0.8109.
%
\end{abstract}


\section{Introduction}

\begin{figure*}
\centering
  \includegraphics[width=0.93\textwidth]{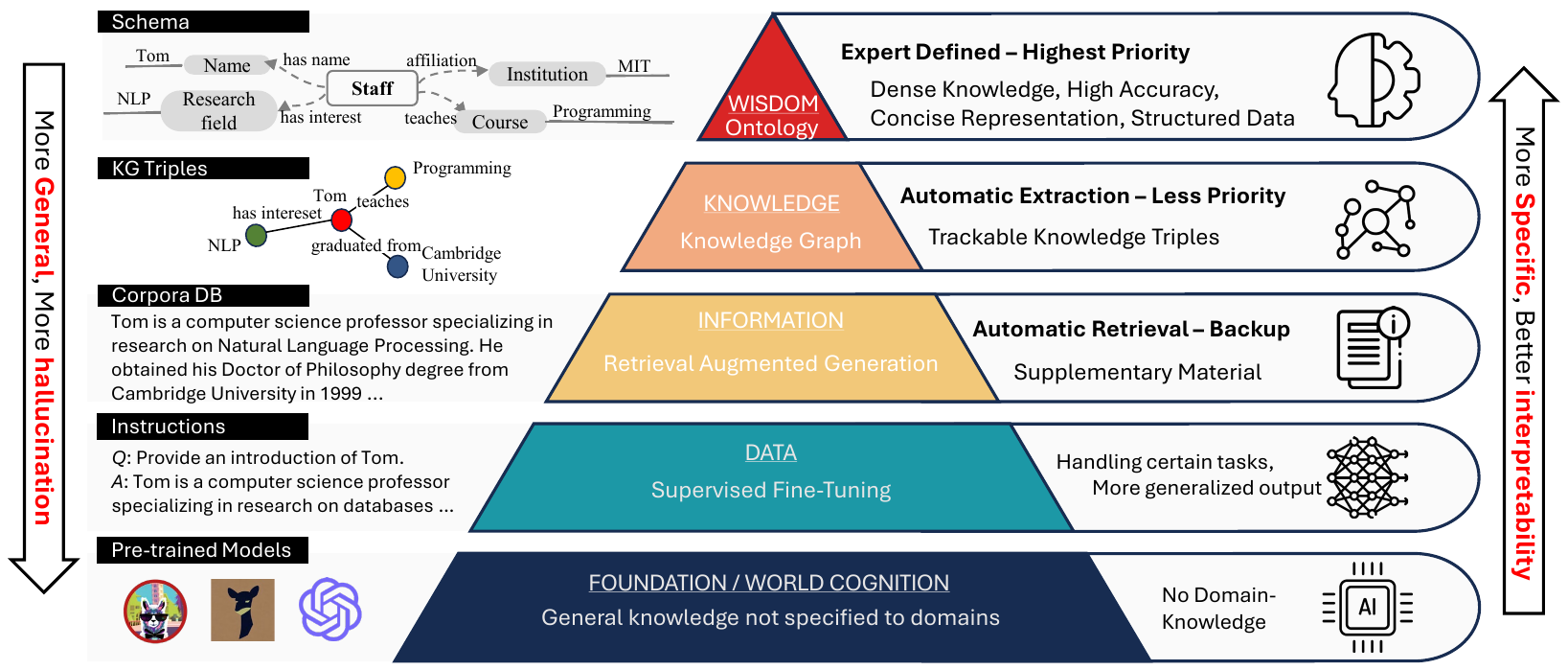}
  \caption{PolyRAG Framework based on the concept of Knowledge Pyramid: different layers signify varying levels of specificity and richness.
The layers are mutually complementary, working together for a balance between recall and precision.}
  \label{fig:teaser}
\end{figure*}

\IEEEPARstart{S}{ignificant} advancements have been made in Large Language Models (LLMs), including proprietary ones like ChatGPT and GPT-4, as well as open-source variants like FLAN \cite{1-3-flan} and LLaMA \cite{1-4-llama}. 
These models have demonstrated remarkable achievements across a wide range of general knowledge tasks such as language comprehension \cite{radford2018openaiGPT, hadi2023llmsurvey}, logical reasoning \cite{kojima2022zero_shot_reasoner, chang2024evaluation_survey}, and complex question-answering \cite{1-5-chatgptReview}.
However, the one-size-fits-all nature of general LLMs fails to meet the specific demands for professional or personalized knowledge, such as law \cite{2-lawyer, 2-chatlaw} and finance domains \cite{4-3-fingpt, 2-fin-pixiu}. 
A straightforward solution is to utilize Supervised Fine-Tuning (SFT) to tailor LLMs to domain-specific tasks \cite{1-2-openaiSFT}. 
%
%
%
Examples include AlpaCare \cite{1-6-alpacare}, Mental LLaMA \cite{1-7-mentalllama} and Zhongjing \cite{2-med-zhongjing} in medical domain, BloombergGPT \cite{1-8-bloomberggpt}, Pixiu \cite{2-fin-pixiu}, and DISC-FinLLM~\cite{4-3-discfinllm} in financial domain, and DISC-LawLLM~\cite{2-disclawllm} in law domain.
Nonetheless, this risks catastrophic forgetting \cite{1-9-forget} of the general knowledge, and is prone to model hallucination \cite{1-10-hallucination}. 

Prevalent alternatives of Retrieval Augmented Generation (RAG) are thus introduced as a means to enhance the domain-specific knowledge comprehension \cite{1-11-lewis2020rag}. 
Rather than solely relying on the generation capabilities of LLMs to produce answers, RAG takes a different approach by incorporating information retrieval techniques. 
It retrieves relevant information from existing resources and utilizes this information to enrich the context of the prompt \cite{2-1-ragreview}. 
This enables in-context learning \cite{1-incontext-survey} or few-shot learning \cite{1-fewshot-survey} and makes the LLMs' output more stable, accurate, traceable, and interpretable. 
However, early implementations of RAG have relied on unstructured textual data chunks, which are often obtained by partitioning each document in the corpora database into segments with a predefined chunk size and thus not organized in a specific way \cite{2-1-naiverag, 2-1-rag}.
To address this limitation, Knowledge Graphs (KGs) have been incorporated into RAG \cite{1-14-kaping, 1-15-kgretrieve,1-kg-recom}.
%
%
The incorporation of KGs in RAG is a logical choice, as KGs have long been recognized as a promising augmentation method for retrieval tasks.
%
%
In fact, in traditional retrieval tasks, another solution called Ontologies has been found to be suitable for domain-specific tasks due to their formally and strictly defined schemas \cite{1-13-onto, 1-ontochatgpt}. 
Nonetheless, the use of Ontologies in RAG has been underexplored, partly because the process of defining the Ontology schema requires significant human effort.

In previous RAG methods, the integrated KGs or Ontologies primarily serve the purpose of incorporating additional domain-specific information, thus enhancing the ``recall'' by increasing the likelihood of locating relevant information.
However, the aspect of ``precision'' has not received specific attention in these methods. For instance, Ontologies are recognized for their precision due to the stringent regulations governing schema and SPARQL syntax. 
Nonetheless, the Ontology search process does not actively participate in the querying process within existing RAG methods.
Figure \ref{fig:teaser} illustrates that different knowledge bases exist at varying levels of specificity, forming a pyramid-like structure. 
The higher layers of the pyramid contain more specific and structured information, providing a greater likelihood of accurately answering domain-specific questions. Hence, these higher layers are considered more ``precision''-friendly.
On the other hand, the lower layers of the pyramid are more inclined towards ``recall''-friendliness. 
They possess more flexible definitions and can store richer information, albeit potentially sacrificing precision.

The observations we have made throughout our research motivate us to develop a multi-layer knowledge pyramid and design a customized querying strategy within the RAG framework. The main objective of this approach is to achieve a better balance between ``precision'' and ``recall'' when retrieving information from heterogeneous knowledge sources. By organizing knowledge into hierarchical layers, we can leverage the strengths of each type of knowledge representation to enhance overall system performance in information retrieval and question-answering tasks. Specifically, we structure the knowledge pyramid with three distinct layers: Ontologies, Knowledge Graphs (KGs), and chunk-based raw text. The Ontology layer provides a highly structured semantic framework with formal definitions and relationships curated by domain experts, enhancing precision. The KG layer captures entities and their interrelations extracted from data, offering a rich semantic network that balances precision and recall. The raw text layer comprises unstructured textual data, providing extensive coverage and improving recall. 

Moreover, during the construction of this pyramid, we employ cross-layer augmentation techniques to ensure comprehensive coverage, dynamically updating the Ontology schema and instances without extensive human intervention. We also maintain compactness by applying cross-layer filtering methods, removing redundant triplets in the KG layer for knowledge condensation. This method enables the layers to interact naturally through the guidance of the distribution in semantic space, which has proven to be effective for creating a more comprehensive and compact knowledge base.

\begin{figure*}[t]
    \centering
    \includegraphics[width=1\textwidth]{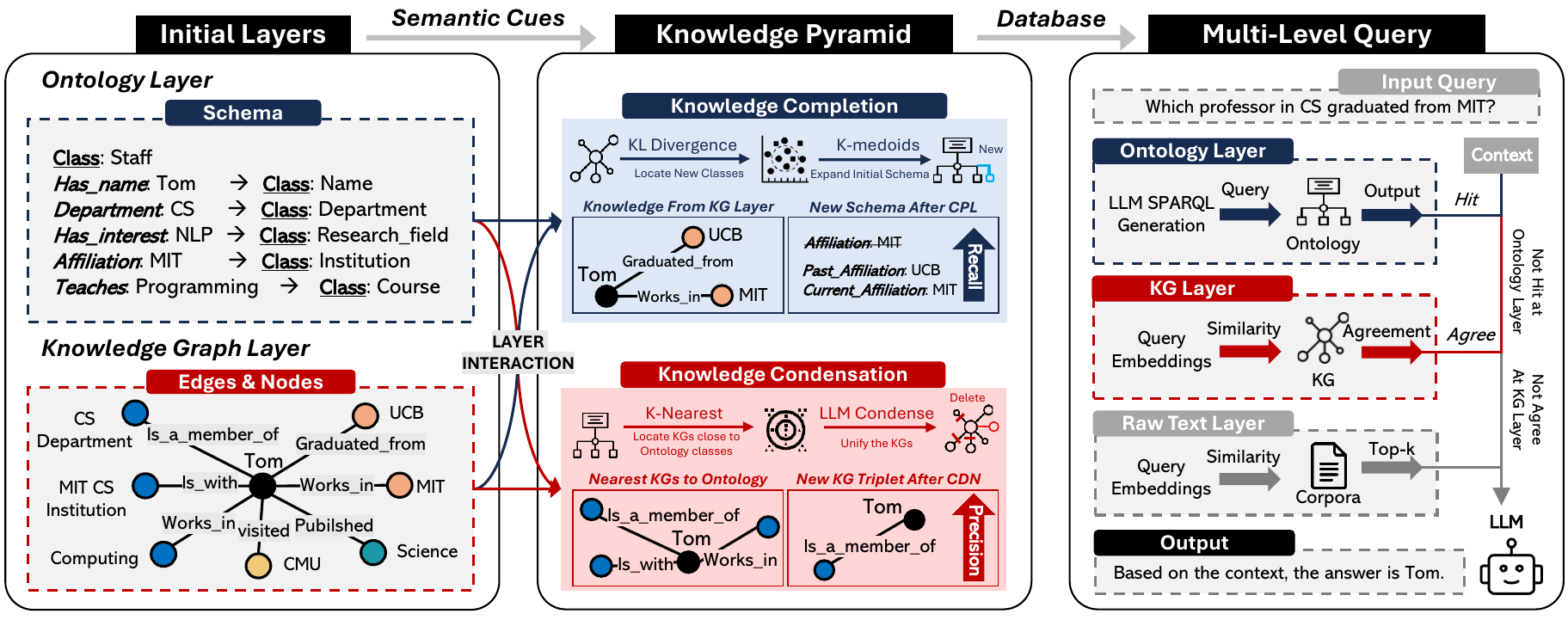}
    \caption{The framework of PolyRAG includes: (1) Knowledge Pyramid Construction, starting from extracting knowledge from corpora data to form the initial layers, then performing the layer interactions, which are knowledge completion and condensation. (2) Given the refined Knowledge Pyramid, the LLM inference obtains the contexts through a multi-level query, which retrieves the knowledge from the higher layer down to the lower layer in a waterfall pattern.}
    \label{fig:framework}
\end{figure*}

The retrieval process follows a waterfall model, starting from the top of the pyramid and progressively moving down until a confident answer is reached. The system first attempts to resolve queries using the most structured and precise knowledge in the Ontology layer. If certainty is not achieved, the search continues to the KG layer, and then to the raw text layer if necessary. This strategy ensures efficient retrieval by prioritizing higher-precision sources while still accessing broader information when needed. We name our approach PolyRAG to reflect its nature of multi-layer knowledge organization and multi-level querying strategy, signifying the incorporation of knowledge from multiple layers and the execution of multi-level querying during the search process. The overall framework of PolyRAG is shown in Figure \ref{fig:framework}.

Furthermore, this paper presents a contribution by introducing two benchmarks for domain-specific knowledge retrieval. 
The first benchmark is specifically designed for the academic domain and has been meticulously constructed by the authors themselves. 
It encompasses extensive information regarding 1,319 staff members, 2,061 courses, 31 departments, as well as classrooms and facilities at XXX University.
In addition, the authors have extended an existing public dataset called FiQA \cite{4-1-fiqa} in the financial domain to create the second benchmark. 
We restructured the knowledge from this dataset into a similar pyramid structure, allowing for the application of the proposed methodology.
Both benchmarks will be made available to the community. 

\section{Related Works}

\subsection{Domain-Specific Large Language Models}
%
%
LLMs have become integral in various applications, such as chatbots, writing assistants, customer service automation \cite{2-app1, 2-app2}.
Domain-specific LLMs are a subset of LLMs that have been tailored to understand and generate text within a particular area of expertise, such as law \cite{2-lawyer, 2-chatlaw, 2-disclawllm}, medicine \cite{1-6-alpacare, 2-med-zhongjing, 1-7-mentalllama}, or finance \cite{2-domainspecific, 4-3-fingpt, 2-fin-pixiu, 4-3-discfinllm}. These models aim to provide higher accuracy and more relevant outputs than general-purpose LLMs when dealing with specialized content \cite{1-2-openaiSFT}.
%
%
A common approach is to adapt instruction fine-tuning based on the general LLMs \cite{1-2-openaiSFT} using domain-specific copra. 
%
%
However, research has shown that the generality of the fine-tuned model will diminish, resulting in catastrophic forgetting \cite{1-9-forget, 2-forget-investigating}. 
To tackle the limitations, the approach of prompting LLMs with extra domain knowledge as contexts is widely adopted.

\subsection{Retrieval-Augmented Generation}
\label{sec:RAG}
Retrieval-Augmented Generation (RAG) enhances the generative capabilities of language models by incorporating retrieved knowledge for in-context learning \cite{2-1-rag,2-1-ragreview}. 
While general language models excel in producing responses for general queries, they are prone to generating hallucinations when tasked with domain-specific knowledge usage. 
RAG addresses this issue by retrieving established knowledge corpora and providing this information as context to the language model \cite{2-1-ragreview}. 
%
%
NaiveRAG represents the most basic architecture within this framework, in which the system retrieves the top-k documents that are most relevant to the query and integrate them into the prompt, thereby grounding the responses in more relevant information \cite{2-1-naiverag}. 

Expanding on NaiveRAG, advanced RAG incorporates additional modules or structures to improve retrieval precision. Reranking is a notable example, where a reranker is employed to refine the initial ranked list (e.g., Re2G \cite{2-1-re2g} and bge-reranker \cite{4-3-bge}, both are based on BERT \cite{2-1-bert}).
Furthermore, studies have indicated that excessive noise and lengthy context can have a negative impact on inference performance. To address this, prompt compression methods such as Selective Context \cite{2-1-selective} and LLMLingua \cite{2-1-lingua} have been developed. 
These methods emphasize key information while reducing noise and context length, as discussed in \cite{2-1-ragreview}.
While current knowledge retrieval methods that rely on single queries can offer some contextual assistance for domain-specific knowledge, they heavily rely on the expressive abilities of the original collections of information and consistently require a trade-off between more detailed knowledge and less noisy information \cite{chongwahngo2005trecvid, chongwahngo2008trecvid}. These techniques do not effectively tackle the issues of retrieving responses that require integrating information from numerous sources, nor do they fulfill the need for dense knowledge in question-and-answer interactions. To address these constraints, we suggest employing varied knowledge representations. Our strategy seeks to optimize the distribution of query duties among knowledge bases in a knowledge pyramid, with the goal of reducing redundancy and increasing the amount and accuracy of information.

\subsection{Knowledge-Augmented Language Models}
The Knowledge-Augmented Language Model approach involves integrating LMs with additional knowledge bases to facilitate in-context learning \cite{Liu2019KnowledgeAugmentedLM}. 
In addition to incorporating knowledge as raw texts \cite{2-1-naiverag}, knowledge graphs (KGs) \cite{KGRAG2023Baek, 3-llmandkg, 1-15-kgretrieve} have gained popularity, and Ontologies \cite{1-13-onto} have also been utilized, albeit less frequently.
In most KG augmentation methods \cite{3-llmandkg}, the RAG framework is followed by replacing the raw text chunks with retrieved KG triplets. 
KAPING \cite{1-14-kaping} serves as an early example of this approach, which has been later refined in RRA \cite{Wu2023RRARetrieveRewriteAnswerAK}.
Regarding Ontologies, instead of being used as an individual knowledge base, they are often employed as assistants for generating KG triplets \cite{3-llmandkg} (e.g., Text2KGBench \cite{mihindukulasooriya2023text2kgbench}) or for augmenting textual corpora (e.g., EKGs \cite{baldazzi2023finetuneontologicalreasoning}, OntoChatGPT \cite{1-ontochatgpt}).
The integration of various forms of knowledge bases poses a significant challenge.

\section{Knowledge Pyramid Construction}
\label{sec:pyramidConstruction}

Let us establish the knowledge pyramid as the foundational base for PolyRAG.
The construction process begins by creating three distinct knowledge banks, each guided or supported by an Ontology, a knowledge graph, and the raw texts, following common practices.
These banks (denoted as $\mathcal{O}$, $\mathcal{K}$, and $\mathcal{T}$, respectively) form the initial layers of the pyramid.
The essence of our proposed methodology lies in fostering interactions between these layers, aimed at enhancing the overall comprehensiveness and compactness of the knowledge base.
%

\subsection{Construction of Initial Layers}
\label{sec:Construction_layers}

\noindent \textbf{Ontology Layer:} 
The Ontology layer $\mathcal{O}=\{\mathcal{O}_s,\mathcal{O}_i\}$ consists of a schema $\mathcal{O}_s$ and corresponding instances $\mathcal{O}_i$.
Defining an Ontology schema typically requires significant time and effort from human experts, and achieving a comprehensive schema can be challenging.
To simplify the process, one approach is to initially extract a sub-domain schema from general Ontologies like WordNet \cite{miller1995wordnet} or ConceptNet \cite{speer2017conceptnet}.
This extracted schema can serve as a starting point and be refined using the semi-automatic approach presented in the Knowledge Completion.
%

With the schema $\mathcal{O}_s$ that we extracted using Web Ontology Language (OWL) \cite{Antoniou2004OWL}, we can guide LLMs to extract instances from the raw text layer $\mathcal{T}$ for each of the class-property pair $(c,a)\in \mathcal{O}_s$, where $c$ is a class (e.g., \textit{staff}) and $a$ is one of its property (e.g., \textit{graduated\_from}).
This can be written as a prompt function
%
\begin{tcolorbox}[boxrule=0pt, frame empty]
    $f_{ins}(c, a;\,p)$: Given a paragraph \{\textit{p}\} from the \{\textit{domain}\} domain, please identify instances of the Ontology relationship where a class \{\textit{c}\} has the property of \{\textit{a}\}. Note that the property may consist of multiple entities.
\end{tcolorbox}
Executing this function results in instances that fulfill the specified relationship. 
By repeatedly applying this function to each paragraph, the set $
\mathcal{O}_i$ is constructed as
\begin{equation}
    \mathcal{O}_i=\{f_{ins}(c, a;\,p)\,\vert\,\forall(c, a)\in \mathcal{O}_s,\,\forall p\in\mathcal{T}\}.
\end{equation}
It is important to note that the specific implementation of the prompt may vary among LLMs.
Additionally, in certain cases, it may be beneficial to provide examples to initiate few-shot learning for extracting high-quality information, depending on the capabilities of the LLMs.
Our implementations are available in the Github repository.

\begin{figure}
    \centering
    \includegraphics[width=0.95\linewidth]{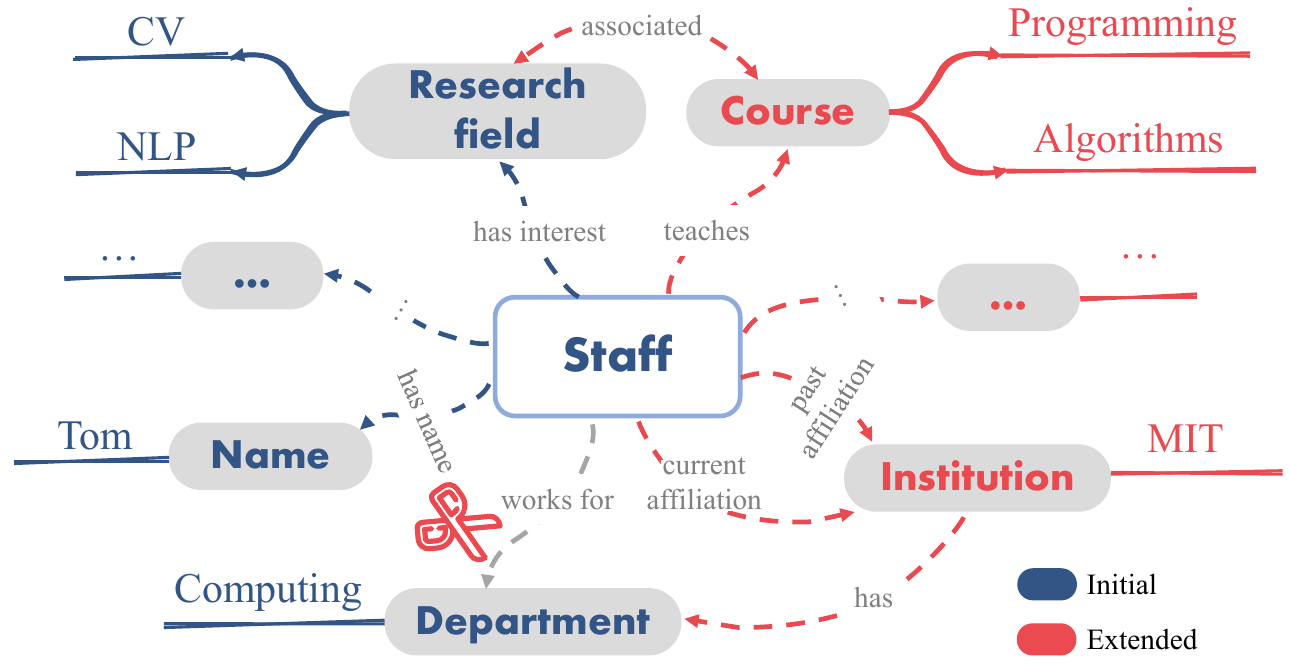}
    \caption{Knowledge completion through the cross-layer interactions. The comprehensiveness of the initial Ontology schema has been improved with the noteworthy concepts identified from the knowledge graph layer.}
    \label{fig:completion}
\end{figure}

\begin{figure*}
    \centering
    \includegraphics[width=0.9\textwidth]{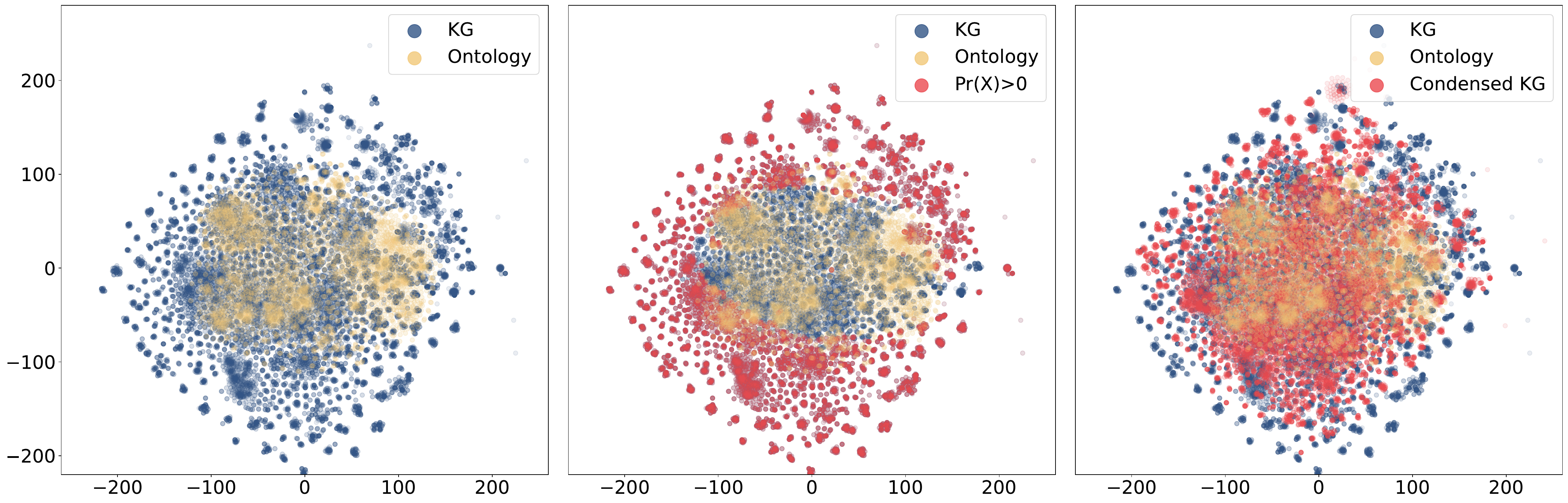}
    \caption{The distribution of KG and Ontology: the original distributions (left), samples with a high priority for knowledge completion are highlighted in red (middle), and  the condensed KG after knowledge condensation (right).}
    \label{fig:clusters}
\end{figure*}

\noindent \textbf{Knowledge Graph Layer:} 
We adopt Open Information Extraction (OpenIE) \cite{3-3-openie} to extract KG triplets from raw texts. 
However, direct extraction often results in noisy output, including irrelevant or duplicate entities. To address this, we draw inspiration from the multi-round prompting approach of LLM2KG \cite{3-3-llm2kg} and reimplement it by introducing four functions: $f_{par}(p)$ for paraphrasing, $f_{ent}()$ for entity extraction, $f_{rel}()$ for relation completion, and $f_{dis}()$ for disambiguation.
These functions form a cascade for KG triplet extraction and refinement as
\begin{equation}
    f_{kg}(p)=f_{dis}\big(f_{ent}(f_{par}(p)),\,f_{rel}(f_{par}(p))\big).
\end{equation}
It constructs the initial knowledge graph layer as
\begin{equation}
    \mathcal{K}=\{f_{kg}(p)\,\vert\,\forall p\in\mathcal{T}\}.
\end{equation}
Specifically, the detailed implementation of four functions is shown as

\begin{tcolorbox}[boxrule=0pt, frame empty]
    $f_{par}(p)$: Determine the factual claims from a given paragraph \{$p$\}. Put these facts into short phrases with basic grammar. Remember that every statement should have a distinct meaning, and pronouns should be avoided.
\end{tcolorbox}
\begin{tcolorbox}[boxrule=0pt, frame empty]
    $f_{ent}(f_{par}(p))$: Extract the noun entities in phrases from the given sentences $f_{par}(p)$. The entities should not contain any comma, and each entity should be unique during extraction.
\end{tcolorbox}
\begin{tcolorbox}[boxrule=0pt, frame empty]
    $f_{rel}(f_{par}(p))$: Given the reference context $f_{par}(p)$ and relevant entities, complete the relations between two entities. Notice that a triple should only contain one entity as head, one verb or verb phrase as relation, and one entity as tail. Separate the head, relation, and tail with a comma.
\end{tcolorbox}
\begin{tcolorbox}[boxrule=0pt, frame empty]
    $f_{dis}(f_{ent}(), f_{rel}())$: You are given several triples $f_{rel}()$ with their entities $f_{ent}()$. These triples consist of subject-predicate-object elements, separated with a comma, but may contain ambiguities or inaccuracies. Your task is to refine and disambiguate these triples to ensure that they accurately reflect the entities and relationships described in their source texts without duplication or omissions. If the relationships have the same semantic meaning, rewrite the triples with the same relation. If the triple has already been mentioned with the same meaning as previous triples, delete it.
\end{tcolorbox}

Therefore, the initial layers, including both ontology and KGs, plus the last raw-text layer, which serves the purpose of backup knowledge in case the retrieval is not accomplished through the preliminary layers, formulate the precondition of the procedure of completion and condensation, thus can operate multi-layer query of PolyRAG framework.

\subsection{Knowledge Completion}
\label{sec:completion}

As shown in the pyramid of Figure~\ref{fig:teaser}, the higher layers are more structured but not easy to define or extract, resulting in limited coverage.
One common issue is the absence of important classes or attributes from the expert-defined Ontology schema, which has historically hindered the ease of defining Ontologies.
By utilizing the pyramid framework, we can address this issue in a data-driven manner.
The idea is to identify noteworthy concepts and relations that exist in the lower layers but are absent from the higher layers. 
These identified elements are then incorporated into the higher layers to enhance knowledge completion.
%
%
For instance, our statistics indicate that while \textit{past\_affiliations} frequently appear in the knowledge graph and raw texts, but are missing from the Ontology. 
This oversight stems from the assumption in the initial Ontology that all facility members are affiliated with the same university, differing only by department, leading to the use of \textit{current\_department}.
However, \textit{past\_affiliations} are valuable for user queries about alumni or consultants from other universities. 
Thus, it would be more effective to replace \textit{current\_department} with separate \textit{past\_affiliations} and \textit{current\_affiliations} to enhance the Ontology's completeness and usability.
These missing noteworthy concepts are indeed indicators for positions where the Ontology and knowledge graph differ significantly in semantic coverage.
Therefore, if we can map the Ontology and knowledge graph into a same semantic space and those concepts can be identified by measuring the divergence of their coverage. 

To achieve this, we begin by transforming class-attribute pairs in the Ontology layer $\mathcal{O}$ into subject+relation phrases that align with the format in the knowledge graph layer $\mathcal{K}$.
Instructor embedding \cite{INSTRUCTOR} of phrases at both layers are then encoded into a common semantic space.
We then learn for each layer a multivariate Gaussian using
\begin{align}
    F(\mathbf{X})&=\frac{1}{(2\pi)^{\frac{n}{2}}\vert\mathbf{\Sigma}^{\frac{1}{2}}\vert}e^{-\frac{1}{2}(\mathbf{X}-\mathbf{\mu})^\top\mathbf{\Sigma}^{-1}(\mathbf{X}-\mathbf{\mu})},
\end{align}
where $n$ is the dimension of the space, $\mathbf{X}$ denotes a phrase embedding, and $\mathbf{\mu}$ and $\mathbf{\Sigma}$ are the mean and covariance matrix, respectively.
The semantic distributions of the Ontology and knowledge graph are modeled as $\mathbf{X}_\mathcal{O}\sim\mathcal{N}(\mathbf{\mu}_\mathcal{O},\mathbf{\Sigma}_\mathcal{O})$ and $\mathbf{X}_\mathcal{K}\sim\mathcal{N}(\mathbf{\mu}_\mathcal{K},\mathbf{\Sigma}_\mathcal{K})$.
The Kullback–Leibler (KL) divergence is then measured on each phrase embedding as
\begin{align}
    D_{\mathbf{X}}(\mathcal{O}\parallel\mathcal{K})&=-F_{\mathcal{O}}(\mathbf{X})\log(F_{\mathcal{K}}(\mathbf{X}))\\
    D_{\mathbf{X}}(\mathcal{K}\parallel\mathcal{O})&=-F_{\mathcal{K}}(\mathbf{X})\log(F_{\mathcal{O}}(\mathbf{X})),
\end{align}
where $D_{\mathbf{X}}(\mathcal{O}\parallel\mathcal{K})$ measures the relative divergence at the point $\mathbf{X}$ by the distribution of $\mathcal{K}$ as the reference, while in $D_{\mathbf{X}}(\mathcal{K}\parallel\mathcal{O})$, the distribution of $\mathcal{K}$ is used as the reference. 

At a location $\mathbf{X}$, we define a function to rate the priority to extract entities from the knowledge graph to the Ontology as
\begin{equation}
    Pr(\mathbf{X})=D_{\mathbf{X}}(\mathcal{O}\parallel\mathcal{K})-D_{\mathbf{X}}(\mathcal{K}\parallel\mathcal{O}).
\end{equation}
The rating outcomes are depicted in Figure \ref{fig:clusters}, illustrating that this function effectively identifies areas where the knowledge graph contains dense information that is comparatively lacking in Ontology.
We then apply k-medoids clustering on knowledge graph triplets within this area to locate new Ontology classes and attributes that can be integrated into the Ontology schema, along with corresponding instances.
This process can be performed iteratively, allowing for continuous refinement and improvement. 
As a result of these interactions between the two layers, the comprehensiveness of the pyramid is progressively enhanced.
Figure~\ref{fig:completion} provides a partial view of the Ontology to illustrate the completion process. 
In addition to the previously mentioned modifications to affiliations, entity Class and their connections to Staff and Research Interests have also been identified.
The inclusion of them are useful for student queries seeking advisors.

\begin{figure*}
    \centering
    \includegraphics[width=0.95\textwidth]{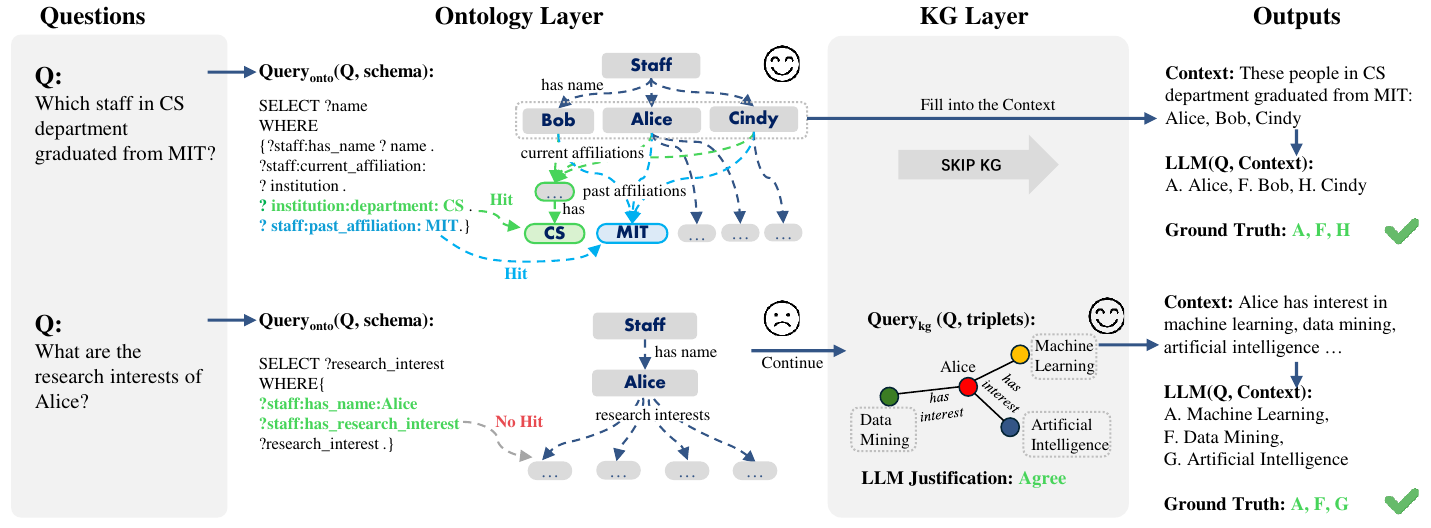}
    \caption{Illustration of multi-level querying: the two queries are addressed at the Ontology and Knowledge Graph layers, respectively, bypassing the lower levels.}
    \label{fig:cases}
\end{figure*}

\subsection{Knowledge Condensation}

The process of Knowledge Completion is designed to enhance the richness and comprehensiveness of information within the knowledge pyramid in a bottom-up manner, which improves the overall ``recall'' when retrieving information using the pyramid framework. By enriching the lower layers, such as raw texts and knowledge graphs, we ensure that more relevant information is available for retrieval.
In addition, we perform Knowledge Condensation to improve the compactness and efficiency of the pyramid in a top-down manner, thereby enhancing ``precision'' in information retrieval. 

The principle behind this approach is to leverage the well-structured knowledge in the higher layers, particularly the Ontology layer, to eliminate redundant or less relevant information present in the lower layers. This is implemented by using each class attribute in the Ontology as an anchor and exploring its neighborhood within the knowledge graph to identify a set of the $k$ nearest knowledge graph triplets ${\mathbf{X_k}}$. By focusing on these closely related triplets, we can filter out extraneous data and refine the knowledge base to include only the most pertinent information. This top-down refinement ensures that the system provides precise answers by relying on well-defined concepts and relationships, ultimately improving the quality and relevance of the retrieved information.
A prompt function is applied to instruct the LLM to summarize those triplets into compact ones as
\begin{tcolorbox}[boxrule=0pt, frame empty]
    $f_{con}(\{\mathbf{X_k}\})$: Please condense the set of knowledge graph triplets $\{\mathbf{X_k}\}$ obtained from the \{\textit{domain}\} domain by eliminating redundant triplets and summarizing the remaining ones in a more concise manner. Here are some examples for your reference: \{\textit{examples}\}. The condensation process should follow the logic of \{\textit{CoT}\}.
\end{tcolorbox}
%

By repeating this condensation process for all anchors in the Ontology layer, we systematically traverse and refine the corresponding neighborhoods in the knowledge graph. This iterative application results in the formation of the condensed knowledge graph layer, as depicted in Figure \ref{fig:clusters}. In this condensed layer, redundant and semantically similar triplets are merged based on their relations to the anchor concepts defined in the Ontology. For example, original triplets such as ``Prof. A works in Z'' and ``Prof. A is with Z institute'', which may have been represented separately in the knowledge graph due to variations in phrasing, are consolidated into a single, more standardized relation of ``\textit{is\_a\_member\_of}''. This condensation not only reduces redundancy but also enhances the semantic clarity of the knowledge representation by unifying similar relationships under a common relation. Such standardization is crucial for maintaining consistency across the knowledge graph and for facilitating more efficient query processing.
\section{Multi-Level Querying}
The pyramid enables PolyRAG in a straightforward top-down querying manner, following the flow Ontology$\rightarrow$Knowledge Graph$\rightarrow$Raw Texts, in which the results are returned if answers are found at higher layers; otherwise, the querying continues to the next layer.
The querying flow is depicted in Algorithm~\ref{alg:PolyRAG}.
Two essential components to build to support the logic are the ways of searching at each layer and conditions to dive next.
The Raw Texts layer can be effectively addressed by employing the NaiveRAG approach.
Regarding the design of the approaches for the remaining two layers, we propose the following strategies.

\begin{algorithm}[t]
\caption{Multi-Level Querying for PolyRAG}
\begin{algorithmic}[1]
    \Function{PolyRAG}{$Q$}
        \State $sparQ \gets Query_{onto}(Q, schema)$ 
        \State $result \gets Onto(sparQ)$
        \If { $result$ is empty}
            \State $triplets \gets Emb{Retrive}_{kg}(Q, \mathcal{K})$
            \State $agreement \gets Query_{kg}(Q, triplets)$
            \If {$agreement$}
                \State $result \gets triplets$
            \Else
                \State $result \gets Emb{Retrive}_{rawtext}(Q, \mathcal{T})$
            \EndIf
        \EndIf
        \State \Return $result$
    \EndFunction
\end{algorithmic}
\label{alg:PolyRAG}
\end{algorithm}

\subsection{Search at Ontology Layer}
At the Ontology Layer, we can utilize SPARQL as the query language. 
SPARQL is a well-defined language that provides precise results when queries are properly structured \cite{1-13-onto}.
To achieve this, we can guide the LLM to generate the SPARQL query by using the following prompt function
\begin{tcolorbox}[boxrule=0pt, frame empty]
$Query_{onto}(Q, schema)$: Given a  question \{$Q$\} within the \{domain\}  domain, please formulate a SPARQL query to retrieve the answer based on the provided Ontology schema. The namespace is \{$schema.base$\}, and the classes are \{$schema.class$\}. The object properties between classes are \{$schema.op$\}, and the classes may also have data properties such as \{$schema.dp$\}. 
\end{tcolorbox}
%

For a query question like ``What are the research interests of Prof. James Wang?'', the LLM extracts the related attributes of ``research\_interest'' with the given $schema$, and generate a SPARQL query to search a staff who satisfies the relevant condition in $\mathcal{O}$. The executing results may include a list of interests or simply be empty due to the lack of knowledge in the Ontology layer.

Figure~\ref{fig:cases} presents an example of a query handled at the Ontology layer. 
The query ``Which staff in CS department graduated from MIT'' is asking for the knowledge of a professor with two restrictions: ``current\_affiliation: CS department'' and ``past\_affiliation: MIT''. This input query is fed into the LLM with the template $Query_{onto}(Q, schema)$. The search process executes SPARQL language to the pre-built ontology database in OWL format. If there exist available results, namely the multi-level querying gets the satisfied context, the process terminates and fills the query results as the knowledge context for question-answering tasks. Due to the accuracy of the knowledge presented at the Ontology layer, the answers seldom fail as long as the knowledge needed is formulated in the ontology database and the LLMs generate the precise SPARQL language for querying. Moreover, since the retrieved results are direct knowledge, which is commonly short phrases of keywords and straightforward to what the query is searching for, it reduced the intelligence requirements of the LLMs, making PolyRAG an easier choice for small and middle-scale LLMs' inference in terms of the context length.  

\subsection{Search at Knowledge Graph Layer}
At the knowledge graph layer, we employ a retrieval approach similar to the embedding-based retrieval utilized in NaiveRAG. 
However, instead of using chunks, we work with triplets.
Treating the triplets as natural language knowledge instead of the graph is being proven to be useful for KG-enhanced LLM question-answering by \cite{1-14-kaping}.
Once the matching triplets are retrieved through the embedding search, we utilize a prompt function to assess the Language Model's agreement on whether the question has been answered adequately as
\begin{tcolorbox}[boxrule=0pt, frame empty]
$Query_{kg}(Q, triplets)$: Given a question \{$Q$\} and the context information provided by the matched triplets from the knowledge graph \{$triplets$\}, please justify whether the provided information is sufficient to accurately answer the question. Respond with either ``Yes'' or ``No'' to provide your justification.
\end{tcolorbox}
%

For example, given a query of ``Which CS staff has interest in cloud computing?'', the LLM may justify a set of triplets as ``agree'' if they include information like ``Prof. A works in CS Department, Prof. A published on cloud computing journal, ... etc.''; inversely, the justification might be ``disagree'' if the triplets do not provide the key knowledge related. The searching will then be processed to the next layer for a more relevant context in order to answer the query.

Figure~\ref{fig:cases} presents an example of a query handled at the Knowledge Graph layer.
The query ``What are the research interests of Alice'' is asking for several research areas, and through the retrieval of the first layer, there is no hit due to the lack of knowledge in the ontology database. Then, the process comes to the second stage, which is to perform the embedding search of top-k similar triplets according to the semantics of the query. Given that a relevant group of triplets is being retrieved, the LLM justification module agrees these triplets should act as the background contexts. Therefore, the LLM is capable of generating the correct answers with the help of precise and concise retrieval.

\begin{table}[t]
\small
\caption{Dataset statistics of the two benchmarks.}
\centering
\renewcommand{\arraystretch}{1.0}
\setlength{\tabcolsep}{13pt}
\resizebox{\linewidth}{!}{
\label{tab:dataset_stats}
\begin{tabular}{lcc}
\toprule
        Dataset                    & \textbf{AcadChall} & \textbf{R-FLUE-FiQA} \\
\midrule
Domain             & Academia        & Finance \\
\#Raw texts              & 5,019                 & 17,072 \\
Ontologies         &               &  \\

\quad  \#Instances   & 25,481  &  225,085\\

\#KG Triples         & 27,920                & 68,737 \\
\#QA pairs           & 512                  & 1,705 \\
QA type                &MCQ\&MAQ              & Open-ended \\
\bottomrule
\end{tabular}
}
\end{table}

\begin{table*}[t]
\centering
\caption{Performance comparison with SOTA methods on the AcadChall and R-FLUE-FiQA datasets.}
\label{tab:results_test}
\resizebox{\linewidth}{!}{
\begin{tabular}{lccc|ccc|cccc}
\bottomrule
&& \textbf{Knowledge}& \textbf{Backend} &\multicolumn{3}{c|}{\textbf{AcadChall}} &\multicolumn{4}{c}{\textbf{R-FLUE-FiQA}} \\\cline{5-11}

\textbf{Group}&\textbf{Method}   &\textbf{Base} & \textbf{LLM}& \textbf{Precision} & \textbf{Recall} & \textbf{F1}& \textbf{BLEU-2} & \textbf{BLEU-4} & \textbf{BERT} & \textbf{HR(\%)} \\
\hline
\multirow{6}{*}{\textbf{Frozen-LLM}} 
&Direct Querying&Pretrained  & LLaMA-7B & 0.2892 & 0.5796 & 0.3470 & 0.4166 & 0.2612 & 0.4658 & 4.4914 \\
 &Direct Querying&Pretrained     & LLaMA2-7B & 0.2400 & 0.4039 & 0.2558 & 0.4106 & 0.2533 & 0.4540 & 4.8735 \\
 &Direct Querying& Pretrained & Vicuna-7B & 0.3195 & 0.6622 & 0.3969 & 0.3552 & 0.2251 & 0.4509 & 4.7271 \\
 &Direct Querying& Pretrained    & Vicuna-13B & 0.3044  & 0.5423 &  0.3606 & 0.4098 & 0.2616 & 0.4933 & 5.6721 \\
  &Direct Querying& Pretrained    & GPT-3.5 & 0.3716  & 0.4031 &  0.3677 & 0.2597 & 0.1525 & 0.5426 & 4.1965 \\
   &Direct Querying& Pretrained    & GPT-4 & 0.1986  & 0.1453 &  0.1636 & 0.4109 & 0.2457 & 0.5415 & 7.6901 \\
\hline

\multirow{3}{*}{\textbf{SFT}}&Finance-LLM &Raw Texts 
  & LLaMA-7B & - & - & - & 0.3990 & 0.2428 & 0.4265 & 3.8447 \\
&Finance-Chat & Raw Texts & LLaMA2-7B & - & - & - & 0.3669 & 0.2265 & 0.4531 & 5.0571 \\
&LoRA & Raw Texts & Vicuna-13B & 0.3234 & 0.6794 & 0.3914 & 0.3768 & 0.2382 & 0.4511 & 4.5854 \\
\hline

\multirow{4}{*}{\textbf{Naive RAG}} &In-Context& Raw Texts  & Vicuna-7B &  0.3621 & 0.8386 & 0.4673 & 0.3159 & 0.2208 & 0.5858 & 7.5769 \\
&In-Context& Raw Texts  & Vicuna-13B & 0.3665  & \textbf{0.9043} &  0.4923 & 0.2934 & 0.1866 & 0.5337 & 5.4326 \\
&In-Context& Raw Texts  & GPT-3.5 & 0.5105  & 0.5791 &  0.5315 & 0.3618 & 0.2147 & 0.5337 & 3.9153 \\
&In-Context& Raw Texts  & GPT-4 & 0.5032 & 0.5094 & 0.4651  & 0.3982 & 0.2404 & 0.5381 & 5.9918 \\
\hline
\multirow{2}{*}{\textbf{Advance RAG}}&Colbertv2 & Raw Texts   & Vicuna-13B & 0.3384 & 0.6693 & 0.3815 & 0.4229 & 0.2979 & 0.5918 & 7.1836 \\
&Bge-reranker & Raw Texts & Vicuna-13B & 0.3003 & 0.6400 & 0.3586 & 0.4400 & 0.3092 & 0.5965 & 7.4097 \\
\hline

\multirow{4}{*}{\textbf{KG Augmentation}}&KAPING &KG  & Vicuna-7B &  0.4365  & 0.8581 & 0.5133 & 0.4079 & 0.2610 & 0.5389 & 5.5468  \\
&KAPING& KG  & Vicuna-13B  & 0.3446  & 0.8945 & 0.4742 & 0.3832 & 0.2403 & 0.5422 & 5.4149 \\
&KAPING& KG  & GPT-3.5  & 0.7326 & 0.6340 & 0.6433 & 0.2599 & 0.1742 & 0.5300 & 4.8785 \\
&KAPING& KG  & GPT-4  & 0.6327 & 0.7065 & 0.6238 & 0.3554 & 0.2319 & 0.5187 & 6.1383 \\\hline

\textbf{\multirow{4}{*}{Ours}} &PolyRAG&Pyramid
 & Vicuna-7B & 0.5306 & 0.8337 & 0.5967  & 0.4485 & \textbf{0.3235} & 0.6008 & 8.6802 \\
&PolyRAG&Pyramid & Vicuna-13B & 0.4330 & 0.8148 & 0.5655 & 0.4444 & 0.3172 & 0.6089 & 8.6954 \\
 & PolyRAG&  Pyramid   & GPT-3.5 & 0.8667  & 0.7860 &  0.8039 & \textbf{0.4459} & 0.3204 & \textbf{0.6114} & \textbf{8.7046} \\
  & PolyRAG&  Pyramid   & GPT-4 & \textbf{0.8711}  & 0.7992 &  \textbf{0.8109} & 0.4002 & 0.2609 & 0.6012 & 7.7705 \\
\toprule
\end{tabular}
}
\end{table*}

\section{Experiments}
\label{sec:experiments}
\subsection{Experimental Setup}
Our experimentation involves two distinct domain-specific benchmarks. 
The first benchmark is Academia Challenge (AcadChall), which has been built by ourselves and focuses on the academic domain. 
It encompasses a comprehensive collection of data obtained from XXX university, including information about 31 departments, 1,319 faculty members, and 2,061 courses. 
Specifically, AcadChall consists of 512 MCQ and MAQ questions that cover topics related to teaching and research. 
These questions are intentionally designed to be more challenging compared to existing benchmarks, as each question presents eight answer choices. 
This design aims to provide a more rigorous assessment of the models' precision.
The second benchmark is FiQA, which is widely recognized and utilized in the finance domain. Our experimentation follows the dataset split provided by FLUE benchmark \cite{4-1-fiqa}, resulting in the creation of the R-FLUE-FiQA dataset comprising 1,705 open-ended questions.
To enhance knowledge completion and condensation, we employ the methods proposed in Knowledge Pyramid Construction Section to construct Ontologies, knowledge graphs, and pyramids based on the two datasets. 
The detailed information about the two benchmarks is shown in Table \ref{tab:dataset_stats}.

We employ instructor-xl \cite{INSTRUCTOR} for embedding and the cosine similarity is used as the metric \cite{cosineSimilarity1991}.
The $k$ values for KG layer and raw texts layer are empirically set to 10 and 5, respectively.

\subsection{Evaluation Metrics}
Our evaluation process encompasses several metrics to assess the performance, including: 
1) For MCQs and MAQs, we employ metrics such as Precision, Recall, and F1 score. 
2) For open-ended questions, we utilize BLEU and BERT similarity, calculated based on MiniLM embeddings \cite{MiniLM2020Wang}.
3) We introduce a novel metric called HitRate ($HR$), which quantifies the proportion of correct entities present in the response.
These metrics collectively allow for a comprehensive assessment of the system's effectiveness across different question types and response formats. 

\subsection{Comparison to SOTA Methods}

We compare PolyRAG with five groups of SOTA methods, including 1) Frozen-LLMs that are pretrained LLMs with frozen parameters and no external context knowledge.
2) SFT (Supervised Fine-Tuning) that are LLMs undergoing supervised fine-tuning on domain-specific datasets.
We include two latest implementations of Finance-LLM and Finance-Chat \cite{financeChatLLM2024Cheng}, both of which are trained on financial corpora. 
%
Additionally, we explore LoRA-based \cite{4-3-lora} SFT.
3) NaiveRAG proposed in \cite{1-11-lewis2020rag}.
4) Advanced RAG that includes two recent models, namely ColBERTv2 \cite{4-3-colbertv2} and Bge-reranker \cite{4-3-bge}.
5) KG-Augmented LLMs that consist of models proposed in \cite{KGRAG2023Baek} that integrate knowledge graphs (KGs) into the LLMs.
We also explored different LLM backbones, including LLaMA \cite{1-4-llama}, LLaMA2 \cite{4-3-llama2}, Vicuna \cite{4-3-vicuna}, Gemini \cite{geminiteam2024gemini}, GPT-3.5 and GPT-4.
By combining them with the five groups and PolyRAG, we conduct a comprehensive set of 23 methods for performance comparison.


The results of our experiments are presented in Table~\ref{tab:results_test}, which showcases the performance comparison between PolyRAG and other state-of-the-art (SOTA) methods across five different groups, various backbone models, and both benchmarks: AcadChall and R-FLUE-FiQA. From the table, it is evident that PolyRAG consistently outperforms other methods in terms of key evaluation metrics such as precision, recall, and F1 scores across all configurations. An encouraging finding from the results is that PolyRAG demonstrates a superior balance between precision and recall compared to other methods. Specifically, among the improvements of PolyRAG over others, the gain in precision (\(22.0\% \pm 17.3\%\)) is more pronounced than that of recall (\(18.4\% \pm 18.6\%\)). This indicates that PolyRAG is particularly effective at accurately retrieving relevant information while minimizing false positives, which is crucial for applications requiring high precision.

Furthermore, with the incorporation of PolyRAG, advanced language models like GPT-4 and GPT-3.5 have achieved significant performance gains. Notably, GPT-4 has achieved a remarkable gain of \(338.6\%\) over its frozen model and \(73.1\%\) over its NaiveRAG run on the AcadChall benchmark. Similarly, GPT-3.5 has achieved gains of \(133.2\%\) over its frozen model and \(69.8\%\) over its NaiveRAG run on the same benchmark.

The notable performance improvements may result from PolyRAG's effective utilization of the hierarchical knowledge pyramid, particularly the higher layers. Our analysis indicates that around \(43.7\%\) (\(38.4\%\)) of questions are addressed at the Ontology layer, and approximately \(42.3\%\) (\(44.8\%\)) are resolved at the KG layer on the AcadChall (R-FLUE-FiQA) benchmark. This suggests that a significant portion of queries benefit from the structured and semantically rich information provided by the Ontology and KG layers. By efficiently leveraging these layers, PolyRAG enhances both precision and recall, leading to significant overall performance gains compared to other methods.

\begin{table*}[t]
\caption{Ablation results on the AcadChall shows the impact of: (1) Completion (CPL) and Condensation (CDN); (2) knowledge layers, including various combinations of Ontology (Ont.), Knowledge Graph (KG), and Raw Text (Raw.).}
\label{tab:ablation}
\centering
\renewcommand{\arraystretch}{1.1}
\setlength{\tabcolsep}{4pt}
\resizebox{\linewidth}{!}{
\begin{tabular}{ll|ccccc|ccccccc}
\bottomrule
\multirow{3}{*}{\textbf{Model}} & \multirow{3}{*}{\textbf{Metric}} & \multicolumn{5}{c|}{\textbf{Knowledge CPL and CND}} & \multicolumn{7}{c}{\textbf{Influence of Knowledge Layers}} \\
\cline{3-14}
& & Naive & \multirow{2}{*}{Baseline} & \multirow{2}{*}{+CPL} & \multirow{2}{*}{+CND} & +CPL & 
 \multirow{2}{*}{Ont.} &  \multirow{2}{*}{KG} &  \multirow{2}{*}{Raw.} & Ont. & KG & Ont. & \multirow{2}{*}{PolyRAG} \\
&& RAG & & & & +CND & & & & +Raw. & +Raw. & +KG & \\

\hline
\multirow{3}{*}{Vicuna-7b}  
& Precision     & 0.3621  & 0.5306  & \textbf{0.6888}  & 0.6557  & 0.5306  
                & 0.4383  & 0.4365  & 0.3621  & 0.4535 & 0.3673  & 0.5251  & \textbf{0.5306}       \\
& Recall        & \textbf{0.8386} & 0.8337  & 0.8325  & 0.7599  & 0.8337  
                & 0.6242  & 0.8581  & 0.8386  & 0.8393 & \textbf{0.8758} & 0.8266  & 0.8337     \\
& F1            & 0.4673  & 0.5734  & 0.5874  & 0.5766  & \textbf{0.5967} 
                & 0.4445  & 0.5133  & 0.4673  & 0.5368 & 0.4840  & 0.5608  & \textbf{0.5967}       \\ 
\hline
\multirow{3}{*}{Vicuna-13b}  
& Precision     & 0.3665  & 0.4230  & 0.4289  & 0.4236  & \textbf{0.4330} 
                & 0.3363  & 0.3446  & 0.3665  & 0.4100 & 0.3747  & 0.4081  & \textbf{0.4330}       \\
& Recall        & \textbf{0.9043} & 0.8848  & 0.8045  & 0.8075  & 0.8148  
                & 0.5811  & 0.8945  & \textbf{0.9043}   & 0.8908 & 0.9050  & 0.8814  & 0.8148     \\
& F1            & 0.4923  & 0.5309  & 0.5595  & 0.5557  & \textbf{0.5655} 
                & 0.3818  & 0.4742  & 0.4923  & 0.5187 & 0.4919  & 0.5166  & \textbf{0.5655}       \\ 
\hline
\multirow{3}{*}{LLaMA2-70b}  
& Precision     & 0.5673  & 0.7661  & 0.8633  & 0.8595  & \textbf{0.8838} 
                & 0.5549  & 0.6666  & 0.5673  & 0.6198 & 0.7421  & 0.6833  & \textbf{0.8838}       \\
& Recall        & 0.3893  & 0.6479  & \textbf{0.6836}   & 0.6445  & 0.6821  
                & 0.4501  & 0.6490  & 0.3893  & 0.5244 & 0.5083  & 0.5980  & \textbf{0.6821}       \\
& F1            & 0.4193  & 0.6678  & 0.7267  & 0.7166  & \textbf{0.7380} 
                & 0.4501  & 0.6080  & 0.4193  & 0.5205 & 0.5505  & 0.5922  & \textbf{0.7380}       \\ 
\hline
\multirow{3}{*}{Gemini-Pro}  
& Precision     & 0.5152  & 0.7956  & 0.8217  & 0.8036  & \textbf{0.8372} 
                & 0.3386  & 0.6388  & 0.5152  & 0.5905 & 0.6890  & 0.7755  & \textbf{0.8372}       \\
& Recall        & 0.3968  & 0.6513  & 0.8597  & \textbf{0.8737}   & 0.8419  
                & 0.3089  & 0.4591  & 0.3968  & 0.4662 & 0.5428  & 0.6697  & \textbf{0.8419}       \\
& F1            & 0.4214  & 0.6899  & 0.7869  & 0.7439  & \textbf{0.8107} 
                & 0.3157  & 0.5099   & 0.4214  & 0.5013 & 0.5821  & 0.6709  & \textbf{0.8107}       \\ 
\hline
\multirow{3}{*}{GPT-3.5}   
& Precision     & 0.5105  & 0.7855  & 0.8044  & 0.7985  & \textbf{0.8667} 
                & 0.4107  & 0.7326  & 0.5105  & 0.6890 & 0.8339  & 0.8567  & \textbf{0.8667}       \\
& Recall        & 0.5791  & 0.7806  & \textbf{0.8152}   & 0.8027  & 0.7860  
                & 0.3945  & 0.6340  & 0.5791  & 0.5818 & 0.6622  & 0.7710  & \textbf{0.7860}       \\
& F1            & 0.5315  & 0.7905  & 0.7939  & 0.7943  & \textbf{0.8039} 
                & 0.3979  & 0.6433  & 0.5315  & 0.6020 & 0.7071  & 0.7939  & \textbf{0.8039}       \\ 
\hline
\multirow{3}{*}{GPT-4}  
& Precision     & 0.5032  & 0.7661  & 0.8063  & 0.8435  & \textbf{0.8711} 
                & 0.3500  & 0.6327  & 0.5032  & 0.6464 & 0.8187  & 0.6083  & \textbf{0.8711}       \\
& Recall        & 0.5094  & \textbf{0.8206}   & 0.7909  & 0.7853  & 0.7992  
                & 0.4369  & 0.7065  & 0.5094  & 0.6779 & 0.7729  & \textbf{0.9032} & 0.7992     \\
& F1            & 0.4651  & 0.7639  & 0.7763  & 0.7873  & \textbf{0.8109} 
                & 0.3677  & 0.6238  & 0.4651  & 0.6303 & 0.7636  & 0.6871  & \textbf{0.8109}       \\
\toprule
\end{tabular}
}
\end{table*}

\subsection{Knowledge Completion and Condensation}
%

We have conducted extensive experiments to evaluate the effectiveness of our proposed Knowledge Completion (CPL) and Knowledge Condensation (CND) techniques by integrating them with a baseline model (PolyRAG without these two modules). The results are presented in the left part of Table~\ref{tab:ablation}, which verifies the effectiveness of our methods across multiple LLM backends, including Vicuna-7b, Vicuna-13b, LLaMA2-70b, Gemini-Pro, GPT-3.5, and GPT-4. Specifically, the baseline model utilizes multi-level querying on the knowledge pyramid without the completion and condensation techniques.

The data in Table~\ref{tab:ablation} clearly demonstrate that both completion and condensation significantly enhance performance. When used individually, CPL yields an average F1 score improvement of $3.8\%\pm 6.1\%$ over the baseline, while CND results in a gain of $2.2\%\pm 3.3\%$. Notably, when both CPL and CND are combined, the performance improvement increases to $6.5\%\pm 6.4\%$, showcasing a more significant boost due to their collaborative effect. Precision is consistently improved across different LLM backends, highlighting the effectiveness of our approach.

In the baseline, around 30\% of questions are answered using more structured data (i.e., Ontology or KG layers). After applying the completion and condensation techniques, this percentage increases significantly to 64\%. For example, with Knowledge Completion, the relation ``\textit{has published in}'', which frequently appears in the KG layer, has been integrated into the Ontology layer. This integration results in questions related to ``publications'' being answered more effectively by the Ontology layer, improving both precision and recall. Similarly, with Knowledge Condensation, there is an increase of around 27\% of questions being resolved at the KG layer. By condensing similar relations such as ``\textit{has published research in}'', ``\textit{has published research articles on}'', ``\textit{has published a paper in}'', and ``\textit{has work published in}'' into unified representations like ``\textit{has published research in}'' or ``\textit{has published research articles in}'', we reduce the number of tokens sent to the LLM for QA. This reduction not only streamlines the context provided to the LLM but also aids smaller LLM backends, such as Vicuna-7B, in understanding the context more effectively, leading to better question-answering results.

Overall, the collaborative use of both techniques results in more significant performance boosts, as evidenced by the highest F1 scores achieved when both CPL and CND are applied together. For instance, in the case of GPT-4, the F1 score increases from 0.7639 (baseline) to 0.8109 with both CPL and CND. These results underscore the complementary benefits of integrating both methods, enhancing both the richness and compactness of the knowledge representation, and ultimately leading to more accurate and efficient question-answering performance across different language models.

\subsection{Influence of Knowledge Layers}
\label{exp:knowledge_layer}
We have conducted a comprehensive examination of the influence of different knowledge layers on various language model backbones. The results, presented in the right part of Table~\ref{tab:ablation}, demonstrate the significant impact that integrating different layers of knowledge has on the performance of language models.

It is evident from the results that the \textbf{Ontology layer} plays a prominent role in enhancing precision across different backbones. For instance, when the Ontology layer is combined with the raw text layer, there is a substantial average precision gain of $25.8\% \pm 7.4\%$. Specifically, for Vicuna-7b, the precision increases from 36.21\% (using raw texts alone) to 45.35\% when combined with the Ontology layer. Similarly, for GPT-3.5, the precision improves from 51.05\% to 68.90\%, and for GPT-4, from 50.32\% to 64.64\%. When the Ontology layer is combined with the KG layer, we observe an average precision gain of $13.7\% \pm 10.3\%$, indicating that the Ontology layer enhances the structured knowledge provided by the KG layer, leading to improved precision.

On the other hand, the \textbf{KG layer} demonstrates a balanced impact on improving both precision and recall. The KG layer alone achieves high recall rates, such as 64.9\% for LLaMA2-70b. This indicates that the KG layer effectively enhances the models' ability to retrieve relevant information due to its rich semantic relationships. Additionally, the KG layer contributes to reasonable precision improvements. For example, for LLaMA2-70b, the precision with the KG layer is 66.66\%, which is higher than using raw texts (56.73\%) or the Ontology layer (55.49\%). Notably, Vicuna-7b and Vicuna-13b, which are among the highest recall scores across all data sources and backbones, tend to select all the choices during the MCQ and MAQ questions and thus lead to an unreal high recall. This analysis verify our claim that balancing precision and recall brings significant meaning to these tasks.

These observations align with our earlier discussions about the benefits of leveraging multiple knowledge layers. The proposed knowledge pyramid significantly improves language model performance by utilizing each layer's complementary strengths. The Ontology layer enhances precision with its structured and precise definitions, while the KG layer contributes to improved recall through its rich semantic connections. The models achieve better overall performance when combined with raw texts, which provide extensive contextual information. These findings confirm that integrating various knowledge layers effectively boosts the capabilities of different backbones, including both smaller models like Vicuna-7b and larger models like GPT-4, thereby validating the effectiveness of our proposed approach.

\subsection{Query Proportion Addressed by Each Knowledge layer}

\begin{figure}[t]
    \centering
    \includegraphics[width=0.7\linewidth]{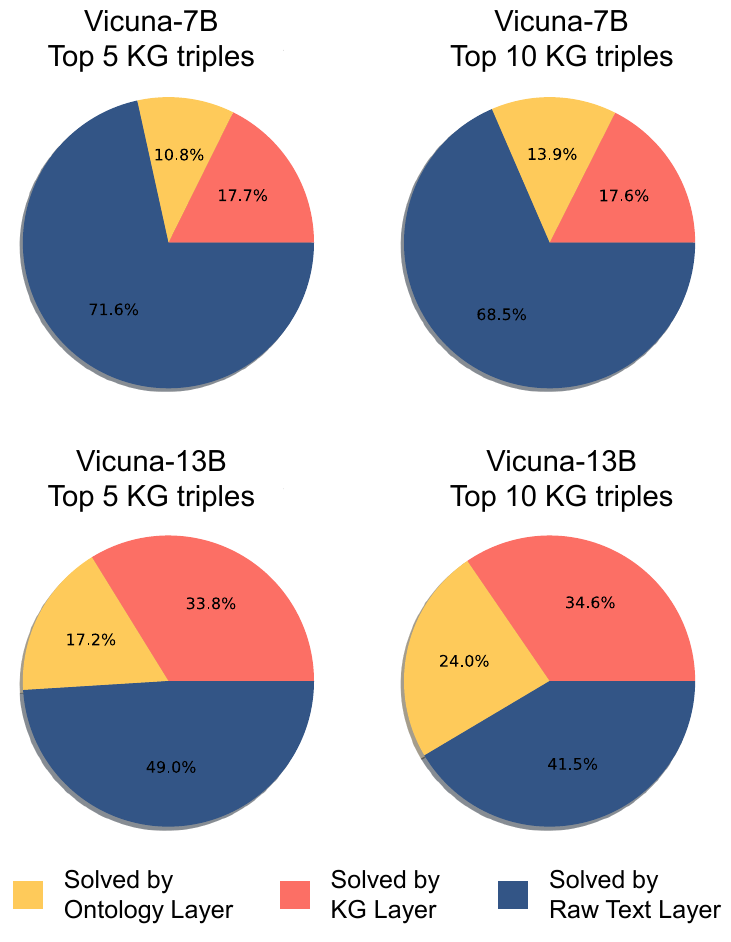}
    \caption{Query proportion addressed by Ontology, KG,  Raw Text Layer on different LLM backbones.}
    \label{fig:topk}
\end{figure}

To gain a deeper understanding of how the PolyRAG framework effectively impacts query outcomes, we conducted an analysis focusing on the proportion of queries addressed by different knowledge layers within the system. Specifically, we visualized the distribution of query resolutions across the various layers of our knowledge pyramid, including the raw text layer, the Knowledge Graph (KG) layer, and the Ontology layer. The results of this analysis are presented in Figure~\ref{fig:topk}, which illustrates how each layer contributes to addressing user queries.

The visualization in Figure~\ref{fig:topk} reveals that a significant portion of queries, at least 28.5\%, can be effectively addressed by the Ontology and KG layers. This finding highlights the crucial role that these structured knowledge layers play in enhancing the performance of the PolyRAG framework. The higher layers provide well-organized and semantically rich information that can directly satisfy user queries without the context for extensive processing. This efficiency not only improves response accuracy but also reduces computational overhead since the water-full framework will terminate when getting available contexts.

Moreover, the proportion of queries addressed by the Ontology and KG layers increases when more KG triplets (i.e., from top-5 to top-10) are included in the context provided to larger Language Model (LLM) backbones. Specifically, we observed that by including five additional KG triplets in the LLM prompt for answering questions, there is a notable increase of 7.6\% (3\%) in the number of queries resolved at the Ontology/KG layers by Vicuna-13b (7b). This demonstrates that enriching the context with more structured knowledge enables the LLMs to leverage this information more effectively, particularly in larger models capable of processing more extensive inputs.

This trend emphasizes the importance of integrating structured knowledge sources to enhance query processing efficiency and effectiveness. The Ontology and KG layers serve as valuable resources that can address a substantial portion of queries directly, reducing the reliance on raw text retrieval and complex language model reasoning. By augmenting the LLM prompts with additional KG triplets, we facilitate better comprehension and utilization of domain-specific knowledge by the language models, resulting in more accurate and contextually appropriate responses.

\section{Conclusion}
This research outlines a preliminary investigation proposing a framework that integrates various knowledge representations to create a knowledge base and offers a multi-level query approach for applying the knowledge pyramid to domain-specific question answering. In this context, we addressed two significant challenges: the problem of prioritizing knowledge in a certain area and the presence of noise in retrieval situations. In order to overcome the limitations, we introduce Knowledge Pyramid multi-level retrieval framework as PolyRAG. It utilizes a sequential retrieval approach and prioritizes knowledge extraction, designed to tackle the challenges posed by the high demand for dense knowledge in domain-specific scenarios and the distractions caused by noisy contexts. The accuracy of PolyRAG has been validated through extensive tests done in AcadChall and R-FLUE-FiQA, surpassing earlier approaches and generating state-of-the-art results.

\bibliographystyle{IEEEtran}




\end{document}